\documentclass[letterpaper, 10 pt, conference]{ieeeconf}  

\IEEEoverridecommandlockouts                              

\usepackage{amsthm} 
\usepackage{amsfonts}
\usepackage{amsmath}
\usepackage{xcolor}

\usepackage{algorithm}
\usepackage[noend]{algpseudocode}

\usepackage{graphicx}
\usepackage[caption=false]{subfig}

\usepackage{cite}

\renewcommand{\baselinestretch}{0.975}
\newtheorem{problem}{Problem}
\title{Topological Trajectory Prediction with Homotopy Classes}
\author{
    Jennifer Wakulicz,
    Ki Myung Brian Lee,
    Teresa Vidal-Calleja,
    Robert Fitch
}

\begin{document}

\maketitle

\begin{abstract}
    Trajectory prediction in a cluttered environment is key to many important robotics tasks such as autonomous navigation. However, there are an infinite number of possible trajectories to consider. To simplify the space of trajectories under consideration, we utilise \emph{homotopy classes} to partition the space into countably many mathematically equivalent classes. All members within a class demonstrate identical high-level motion with respect to the environment, i.e., travelling above or below an obstacle. This allows high-level prediction of a trajectory in terms of a sparse label identifying its homotopy class. We therefore present a light-weight learning framework based on variable-order Markov processes to learn and predict homotopy classes and thus high-level agent motion. By informing a Gaussian Mixture Model (GMM) with our homotopy class predictions, we see great improvements in low-level trajectory prediction compared to a naive GMM on a real dataset.  
\end{abstract}

\section{Introduction}
Understanding the dynamics of mobile agents is important for robots to operate autonomously in environments with pedestrians, vehicles, animals, or other moving entities.
Autonomous driving, service robot navigation through crowds, and similar applications all rely on trajectory prediction to help avoid dynamic obstacles.
Predicting trajectories is also an important form of \emph{intention inference}, where the intention to move in a certain direction or to a given location is inferred~\cite{best_bayesian_2015,Rhinehart_2019_ICCV,eiffert_probabilistic_2020,kavindie_2021}. Examples involving pedestrian trajectories include goal-directed inference~\cite{best_bayesian_2015,Rhinehart_2019_ICCV}, modelling social interactions~\cite{eiffert_probabilistic_2020}, and industrial applications with cobots~\cite{dinh_cobot_2015,cobot_overview}.

We are interested in leveraging environmental context to predict the abstract motion of an agent.
Obstacles in the environment, for example, place constraints on the trajectories we wish to predict.
The environment, therefore, is a rich source of information for trajectory prediction that can be seen to split the set of all possible trajectories into equivalence classes. These equivalence classes summarise the agent's high-level motion in a sparse identifier, sometimes referred to as an \emph{$h$-signature}, that specifies properties such as whether the trajectory passes above or below a certain obstacle, and how many times it does so. 
Having this identifier is powerful, as it gives a general idea of the motion in a sparse representation that can be used as the basis for developing efficient computational frameworks.
For example, $h$-signatures have been used in various path planning contexts~\cite{bhattacharya_search-based_2010, bhattacharya_persistent_2015, bhattacharya_topological_2015,mccammon_topological_2021}.

In trajectory prediction, topological concepts like $h$-signatures allow the extraction of salient, predictive features of trajectories that arise from obstacles or other features of the environment.
Existing trajectory prediction methods, however, typically are based on a geometric representation~(e.g., \cite{zhi_kernel_2020,kavindie_2021}) and can be computationally prohibitive for long-term prediction.
These methods operate on full trajectories and consider all differences between them, including minor differences.
The number of small and insignificant variations, even between topologically equivalent trajectories, can easily become computationally overwhelming.
These geometric algorithms thus trade-off prediction horizon with computational efficiency~\cite{kavindie_2021,Rhinehart_2019_ICCV,eiffert_path_2020,zhi_kernel_2020}.

In this paper, we propose a new topological approach to trajectory prediction by leveraging the salient features of paths summarised as by $h$-signatures. $h$-signatures are both compact in size and provide global information about a trajectory as they only encode how a trajectory moves past an obstacle as opposed to an overall trajectory. The use of $h$-signatures in trajectory prediction can therefore circumvent the aforementioned trade-off that other methods face.

We thus present our method for topology-informed trajectory prediction, which introduces the notion of \emph{partial $h$-signatures}, defined for incomplete trajectories. The concept of partial $h$-signatures decomposes trajectory prediction into high-level and low-level prediction problems.
As initial solutions to these problems, we present a high-level prediction algorithm using VOMP~\cite{ron_power_1993} and a low-level prediction algorithm based on a hierarchical GMM. We demonstrate the behaviour of our method with a simple synthetic dataset, and then extensively evaluate its performance compared to a baseline GMM method without topological knowledge using the ATC shopping mall dataset~\cite{atc_dataset}.
Results show that our method performs as well or better in terms of geometric error, and considerably better in identifying the correct high-level motion.
This indicates that $h$-signatures are a meaningful high-level representation for predicting agent trajectories, thus useful for applications such as target search~\cite{wakulicz_active_2021}.

\begin{figure}[t!]
    \centering
    \includegraphics[width=0.99\columnwidth]{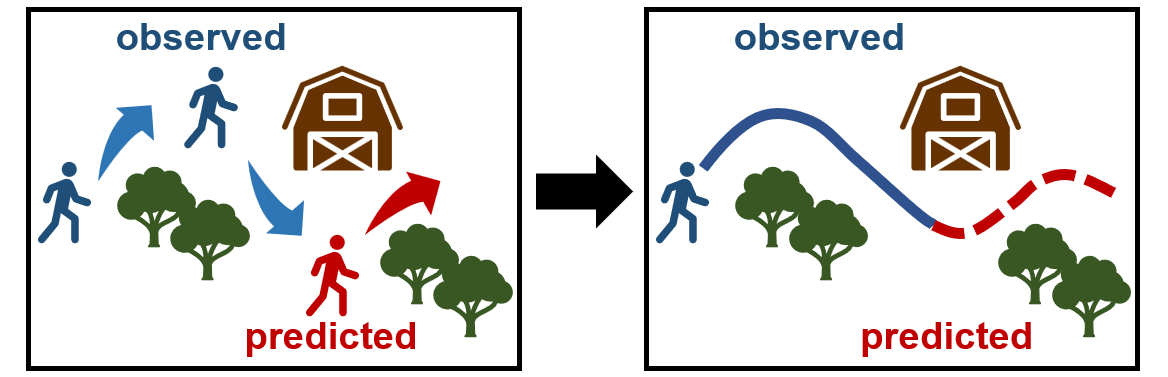}
    \caption{Illustration of the problem considered in this paper. The task is to predict high-level motions then use them to inform lower level trajectory predictions.}
    \vspace{-3.5ex}
    \label{fig:prob_form}    
\end{figure}




\section{Related Work}

Trajectory prediction is a well-studied problem in robotics. Model-based approaches use an estimated or learned dynamic model of an agent's motion to predict an agent's state forward in time~\cite{zhi_spatiotemporal_2019,wakulicz_active_2021}. Model-free learning based methods can be used where dynamics are cumbersome to model. For example, by treating trajectory prediction as a sequence generation problem, recurring neural networks (RNNs) and long-short term memory networks (LSTMs) have been implemented in various contexts with large success~\cite{altche_lstm_2017, shi_lstm-based_2018, eiffert_path_2020, bi_large-scale_2022}. Other approaches have introduced mixture models such as Gaussian mixture models (GMMs) and the kernel trajectory map~\cite{zhi_kernel_2020} to capture the multi-modal nature of an agent's possible future paths, showing greater prediction accuracy over single-mode approaches~\cite{wiest_probabilistic_2012, Ivanovic_2019_ICCV, zyner_naturalistic_2020}.

Beyond removing the burden of acquiring a dynamic model, the power of model-free learning methods lie in their ability to incorporate contextual information into the model. For example, social interactions between agents have been captured in various architectures to improve pedestrian trajectory prediction in crowded environments~\cite{eiffert_probabilistic_2020, vemula_social_2018, alahi_social_2016, Ivanovic_2019_ICCV}. Where crowds are extremely dense and interactions are complex, bulk properties of crowds have been learned instead~\cite{kiss_probabilistic_2021,kiss_constrained_2022}. Even the intent of an agent may be learned, and has been proposed as a powerful contextual cue for long-term trajectory prediction problems~\cite{best_bayesian_2015,yao_bitrap_2021,eiffert_probabilistic_2020,kavindie_2021}. 

Contextual information common and relevant to all trajectory prediction  scenarios is the environment within which the agent is travelling. Obstacles, unreachable regions and road structures -- the \textit{topological features} of an environment -- all dictate how an agent can move through an environment. It seems then that topological information should be a fruitful contextual cue for trajectory prediction. Indeed, topology-aware learning methods have been introduced for trajectory prediction, outperforming a GMM-based approach that does not take advantage of this contextual cue~\cite{pokorny_topological_2016,frederico_carvalho_long-term_2019}.

\section{Background and problem formulation}
\subsection{Homotopy Theory in Robotics}\label{sec:homotopy_theory}
\begin{figure}[t]
    \centering
    \includegraphics[width=0.5\columnwidth]{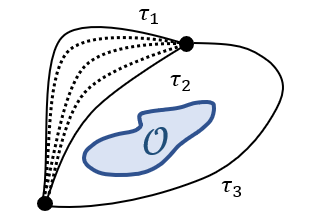}
    \caption{Illustrative example of paths of equal ($\tau_1$, $\tau_2$) and differing ($\tau_3$) homotopy classes.\label{fig:homotopy}}
    \vspace{-1.5em}
\end{figure}

Two paths $\tau_1$, $\tau_2$ in a topological space $\mathcal{D}$ with common start and ending points are \textit{homotopic} if there exists a continuous transformation or deformation from one to the other~\cite{hatcher_algebraic_2002}. Sets of paths homotopic to one another are named \textit{homotopy classes}. Non-trivial homotopy classes arise as a result of obstacles in the space, as deforming some paths into others would require moving through an obstacle and breaking the continuity requirement. This notion is depicted in Fig.~\ref{fig:homotopy}. Paths $\tau_1$ and $\tau_2$ are homotopic as they can be continuously deformed into each other, indicated by the dotted paths between them. These paths are however not homotopic to $\tau_3$, as any deformation into $\tau_3$ would require moving through the obstacle $\mathcal{O}$.

In the context of robotics, obstacles in the environment split the space of an agent's possible trajectories between points from a single homotopy class into a countable number of homotopy classes. The obstacles therefore dictate the number of unique ways an agent may travel through the space from one point to another. For example, in Fig.~\ref{fig:homotopy} the obstacle $\mathcal{O}$ splits the space of trajectories into those that move `above' the obstacle, those that move `below', and those that wind around the obstacle any number of times before moving to the end point. Such abstraction of the high-level motions available to an agent is a powerful tool for navigation, prediction and tracking tasks often encountered in robotics~\cite{bhattacharya_search-based_2010, bhattacharya_persistent_2015, pokorny_topological_2016, frederico_carvalho_long-term_2019, mccammon_topological_2021}. 

\subsection{$h$-signatures as Homotopy Invariants}\label{sec:background:invariants}
To identify which homotopy class a trajectory $\tau$ belongs to, one must compute a \textit{homotopy invariant} -- a unique identifier $h(\tau)$ of a trajectory's homotopy class such that $h(\tau_1) = h(\tau_2)$ if and only if $\tau_1$ and $\tau_2$ are homotopic. There are many ways to construct a homotopy invariant. The most simple and commonly used is detailed in~\cite{hatcher_algebraic_2002} and coined the \textit{$h$-signature} of a trajectory in~\cite{bhattacharya_topological_2015}. Non-intersecting rays are drawn from within each obstacle to the boundary of the environment. These rays are commonly taken to be parallel and emanate upwards from the centres of obstacles. Then to compute the $h$-signature of path $\tau$, a `word' is constructed by following the path and appending letter `$n$' if $\tau$ crosses the ray corresponding to the $n$-th obstacle from left to right, and the letter `$-n$' if $\tau$ crosses from right to left. The final word may then be \textit{reduced} by cancelling all consecutive appearances of $n$ and $-n$. For example, the $h$-signature $(1,2,-2,3)$ reduces to $(1,3)$.

\subsection{Problem Formulation}
Consider an agent traversing through a planar environment $\mathcal{D}$ containing $n$ obstacles $O = \{\mathcal{O}_1,\ldots,\mathcal{O}_{n}\}$.
The agents trajectory is denoted $\mathbf{X} = \{ \mathbf{x}_{1}, \ldots \mathbf{x}_{T} \}$.
We assume that the agent begins at a start location on the boundary $\delta\mathcal{D}$ of the environment and travels to an end location on $\delta\mathcal{D}$. We are given a dataset of $K$ historical, fully observed trajectories $\mathcal{X} = \{X_{\text{obs}}^1,\ldots,X_{\text{obs}}^K\}$, with full knowledge of obstacles. 
Further, in the online setting, we have a partial noisy measurement of trajectory $\mathbf{Y}_{\text{obs}} = \{\mathbf{y}_1,\ldots,\mathbf{y}_{T_{\mathrm{obs}}}\}$ up to time $T_{\mathrm{obs}}$, with a known sensor model $P(\mathbf{y}_{t} \mid \mathbf{x}_{t})$. 

We are then interested in predicting the agent's high-level motion through the environment.
An agent's `high-level motion' can be abstracted from its trajectory in many different ways. For example, one might learn trajectory clusters with a GMM and treat each cluster as a unique high-level motion. 
We instead propose that high-level motion is best abstracted by the trajectory's homotopy class and thus that prediction of a trajectory's $h$-signature is the most appropriate solution. Formally, the problem is:
\begin{problem} (High-level prediction)\label{prob:high-level} Given the partial, noisy measurements of the trajectory $\mathbf{Y}_{\mathrm{obs}}$, predict the $h$-signature $\hat{h}$ of the robot's future full trajectory.
\end{problem}

While the solution to Problem~\ref{prob:high-level} can be used as a tool in many robotics problems, here we are interested in how the $h$-signature may be used to produce a topology-informed low-level prediction for a trajectory:

\begin{problem} (Low-level prediction)\label{prob:low-level} Given the predicted $h$-signature $\hat{h}$ associated with an agent's partially observed measurements $\mathbf{Y}_{\mathrm{obs}}$, predict the full trajectory $\mathbf{X}$.
\end{problem}

\section{Topology-informed trajectory prediction}

\subsection{Overview}\label{sec:overview}
Problems~\ref{prob:high-level} and~\ref{prob:low-level} are challenging to solve directly without modification, as illustrated in the factor graphs shown in Fig.~\ref{fig:pgm}.
In a pedantic Bayesian formulation~(Fig.~\ref{fig:pgm}~\subref{fig:pgm:naive}), one would first predict the underlying trajectory for the entire duration, and subsequently predict the corresponding $h$-signature.
This is because the measurements are conditionally independent of the $h$-signature given the trajectory, and the only given relationship between the trajectory and the $h$-signature is the computation process outlined in Sec.~\ref{sec:background:invariants}. 
In other words, in this view, low-level trajectory prediction precedes high-level prediction, limiting its effectiveness.

Instead, we propose to circumvent the low-level prediction by introducing the notion of partial $h$-signature (green lines, Fig.~\ref{fig:pgm}~\subref{fig:pgm:ours}).
Unlike the usual $h$-signature, partial $h$-signatures can be obtained from an incomplete trajectory, as we detail in Sec.~\ref{sec:partial}.
Given the partial $h$-signature, we predict the full $h$-signature. 
This is achieved with a variable-order Markov process (VOMP) model (red line, Fig.~\ref{fig:pgm}~\subref{fig:pgm:ours}) trained on a dataset (Sec.~\ref{sec:vomp}).
The full $h$-signature can then be used to predict the low-level trajectory, with a model learnt using the historical dataset.
As we readily have access to trajectories and their associated $h$-signature, this can be as simple as learning a mixture of experts for low-level trajectory prediction within each homotopy class.  
To this end, we demonstrate the use of a hierarchical Gaussian mixture model (GMM) in Sec.~\ref{sec:gmm}.

\subsection{Partial $h$-signatures}\label{sec:partial}

As described in Sec.~\ref{sec:homotopy_theory}, homotopy classes exist only for paths between two fixed start and end points. In the context of this paper trajectories are assumed to start and end on different boundary points of the environment. Similar to~\cite{mccammon_topological_2021}, we ensure our description of homotopy classes is valid by applying a quotient map, mapping all boundary points to a single quotient point while preserving the topology of the space. Then, the set of all homotopy classes here is over paths between the quotient point and itself.

With this in mind, the notion of a \textit{partial} $h$-signature is ill-defined in a topological sense. Much like a full $h$-signature, it is calculated by constructing a `word' according to a path's ray crossings. However, it is extracted from an incomplete trajectory, i.e. one that has not yet returned to the quotient point. It is thus crucial to note that a partial $h$-signature is not an identifier of a homotopy class but rather a \textit{predictor}.

To predict full $h$-signatures from partial ones, a notion of compatibility between the two is needed. For a given partial $h$-signature $p$, compatible full $h$-signatures are those whose prefix is $p$. In other words, the set of all $h$-signatures compatible with $p$ is $\mathcal{H}(p) = \{h \mid \exists p', h = pp^{\prime}  \}$.
\subsection{High-level Prediction using Topological Variable Order Markov Processes}\label{sec:vomp}

\begin{figure}[t!]
    \centering
    \subfloat[Pedantic model.]{\includegraphics[width=0.73\columnwidth]{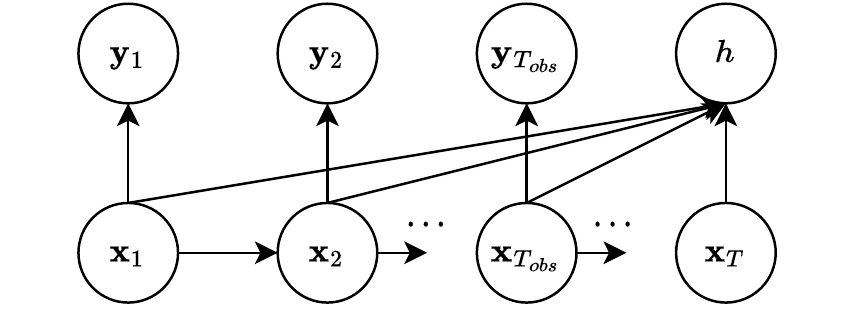}\label{fig:pgm:naive}} \\
    \subfloat[Proposed model.]{\includegraphics[width=0.73\columnwidth]{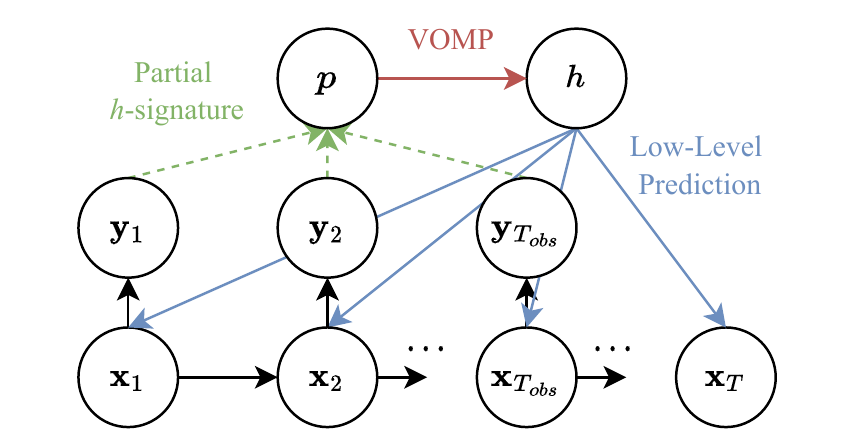}\label{fig:pgm:ours}}    
    \caption{Probabilistic graphical models depicting the problem. Directed arrows from $A$ to $B$ imply availability of a model of $B$ given $A$. Dashed arrows are ignored during inference.}
    \vspace{-1.5em}
    \label{fig:pgm}
\end{figure}

As $h$-signatures in an environment with $n$ obstacles are simply `words' constructed from an alphabet $\mathcal{A} = \{1,..,n,-1,...,-n\}$, prediction of a full $h$-signature given partial $h$-signature can be viewed as a sequence completion problem. The sparse nature of the $h$-signature as a representation of high-level motion allows for relatively simple techniques to be used for sequence generation. We propose a VOMP~\cite{ron_power_1993} for this purpose. Like LSTMs, VOMPs can learn dependencies in data that are of varying length. However, VOMPs are capable of producing probabilistic predictions where LSTMs are not. This is crucial for our approach, as a probability distribution over full $h$-signatures is required to produce probabilistic low-level predictions (blue lines, Fig.~\ref{fig:pgm}~\subref{fig:pgm:ours}).
 
We represent a VOMP with a probabilistic suffix automaton (PSA) as in~\cite{ron_power_1993}. 
Here, a PSA state is a partial $h$-signature $p$ of up to some maximum length $L>0$ constructed from alphabet $\mathcal{A}$ as $p = a_{1}\ldots a_{l}$, $0 \leq l \leq L$. Transitions between two states $p$ and $p^{\prime}$ are allowed only if there exists some $a \in \mathcal{A}$ such that $p^{\prime}$ is a \textit{suffix} of $ap$. Allowed transitions have associated with them a probability that the transition will occur. Transition probabilities learned offline can be used to produce probability distributions over future states online. An example PSA is drawn in Fig.~\ref{fig:vmmp}~\subref{fig:vmmp:automata}.

We follow~\cite{ron_power_1993} closely to train a VOMP over $h$-signatures offline, and adapt their online prediction process to better suit the prediction of $h$-signatures. Our algorithms for online prediction and offline learning are detailed below.

\subsubsection{Online Prediction}
The trained VOMP outputs probabilities of arbitrarily long $h$-signatures. To find the probability of any $h$-signature $P(h)$ one simply takes a corresponding walk through the PSA, multiplying transition probabilities from state to state. However, we are interested in finding the conditional distribution $P(h \mid p)$ over possible full $h$-signatures $h$ given a partial $h$-signature $p$ extracted from a partially observed trajectory. Letting $\mathcal{H}(p)$ be the set of all full $h$-signatures compatible with $p$, the conditional probability $P(h \mid p)$ of any $h \in \mathcal{H}(p)$ can be calculated in typical fashion,

\begin{equation}\label{eq:naive_conditioning}
    P(h \mid p) = \frac{P(h)}{\sum_{h \in \mathcal{H}(p)} P(h) }.    
\end{equation}

However, the set $\mathcal{H}(p)$ of compatible $h$-signatures is of infinite size. Just as agents may walk paths of varying lengths through an environment, the lengths of compatible full $h$-signatures varies. For example, if an agent is observed passing from left to right above the first obstacle in an environment their partial $h$-signature is $(1)$. Later, the agent may or may not pass above any other obstacle. Then, their full $h$-signature may be $(1)$ or $(1,\ldots)$. 

To handle these nuances, we assume that the longest possible $h$-signature an agent will take through the environment is the maximum length $h$-signature present in training data. That is, for any $h \in \mathcal{H}(p)$, $P(h) = 0$ if $|h| > \max(|p|)$. Then, we weigh $h$-signature probabilities in Eqn.~\ref{eq:naive_conditioning} by the probability of observing an $h$-signature of that length in data. Thus, the VOMP is queried in order to make the adjusted calculation

\begin{equation}
    P(h \mid p) = \frac{P(h) \cdot P(|h|)}{\sum_{h \in \mathcal{H}(p)} P(h) \cdot P(|h|) }.
\end{equation}

\subsubsection{Offline Learning}

\begin{algorithm}[t!]
	\caption{PSA offline learning} \label{alg:offline}
	\textbf{Inputs:} $\epsilon$, $L$, $\mathcal{A}$, $h$-signature data \\
	\textbf{Output:} Trained PSA
	\begin{algorithmic}[1]
	    \State initialise tree $T$
	    \State $\mathcal{P} \gets \{a \mid a \in \mathcal{A}, P(a) \geq \epsilon\}$
	    \While{$\mathcal{P}$ not empty}:
	        \State $p \gets \mathcal{P}$.pop()
	        \If{$\mathcal{E}(p, \mathrm{suffix}(p)) \geq \epsilon$}:
	            \State add path to $p$ to $T$
	        \EndIf
	        \If{$|p| \leq L$}:
	            \State $\mathcal{P} \gets \mathcal{P}\cup\{ap \mid a \in \mathcal{A}, P(ap) \geq \epsilon \}$
	        \EndIf
	    \EndWhile
	    \For{all leaves $r$ in $T$}:
	        \If{longest $\mathrm{prefix}(r)$ not in $T$}:
	            \State add path to $r$ to $T$ 
	        \EndIf
	    \EndFor
	    \State PSA $\gets$ leaves of $T$
	\end{algorithmic} 
\end{algorithm}

\begin{figure}[t!]
    \centering
    \subfloat[Prediction suffix tree.]{\includegraphics[width=0.49\linewidth]{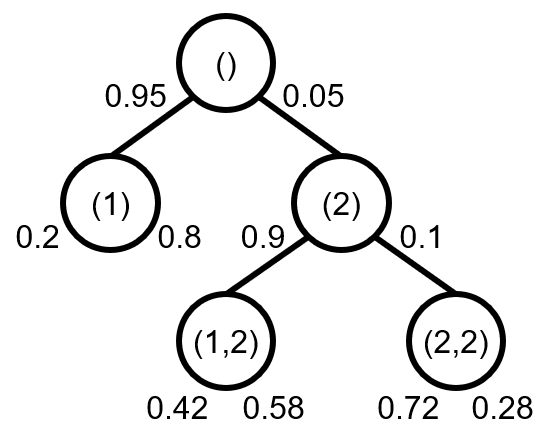}\label{fig:vmmp:tree}}
    \subfloat[Probabilistic suffix automaton.]{\includegraphics[width=0.49\linewidth]{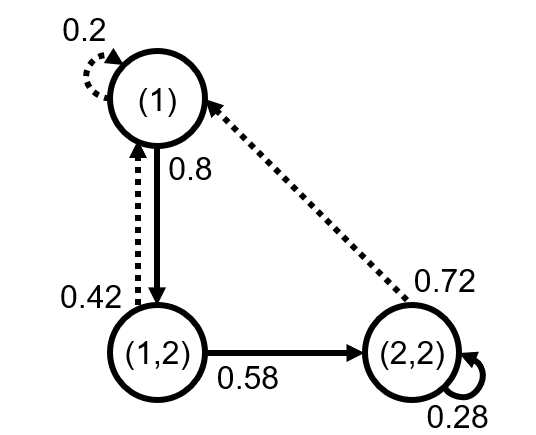}\label{fig:vmmp:automata}}
    \caption{Equivalent methods of representing a VOMP. Values along edges between states are transition probabilities.}\label{fig:vmmp}
    \vspace{-1.5em}
\end{figure}

To facilitate online prediction of full $h$-signatures given partial $h$-signatures, a prediction suffix tree (PST) (Fig.~\ref{fig:vmmp}~\subref{fig:vmmp:tree})  is built over the alphabet $\mathcal{A}$ admitted by all possible $h$-signatures given the environment. The PST is then converted to a PSA (Fig~\ref{fig:vmmp}~\subref{fig:vmmp:automata}) as per~\cite{ron_power_1993}. A high-level overview of this process is provided in Alg.~\ref{alg:offline}.

To construct the PST, it is first initialised with root node corresponding to the `empty' $h$-signature labelled $()$. Paths to suffixes are then successively added to the tree if the suffix has sufficiently strong predictive power. Specifically, a child node labelled with partial $h$-signature $ap$ is added to parent node $p$ if some measure of statistical difference $\mathcal{E}$ between $P(\cdot \mid p)$ and $P(\cdot \mid ap)$ is above a user-defined threshold $\epsilon$. The metric used is the KL divergence scaled by the probability of observing $ap$,
\begin{equation}
    \mathcal{E}(ap, p) = P(ap) D_{KL}(P(\cdot \mid ap) || P(\cdot \mid p)).
\end{equation}
This scaling factor $P(ap)$ serves to avoid the addition of suffixes that have very low probability of occurring, yet give large KL divergence. Suffixes up to length $L$ are tested for their predictive power and added.

To calculate $P(a\mid p)$ and $P(p)$ from training data, Laplace's rule of succession is used. Denoting $\mathcal{F}(p)$ as the frequency with which partial $h$-signature $p$ appears in data, $\mathcal{F}(p^{\mathrm{C}})$ is then used to denote the frequency of the complement event; the frequency of observing any other partial $h$-signature of length $|p|$. Then, 

\begin{equation}
        P(p) \approx \frac{\mathcal{F}(p) + 1}{\mathcal{F}(p^{\mathrm{C}}) + |\mathcal{A}|}.
\end{equation}

Similarly, denoting the frequency with which letter $a$ follows $p$ in observations by $\mathcal{F}(a|p)$ and the frequency of observing any other letter after $p$ by $\mathcal{F}(a^{\mathrm{C}}\mid p)$,

\begin{equation}\label{eq:trans_prob}
    P(a \mid p) \approx \frac{\mathcal{F}(a|p)+1}{\mathcal{F}(a^{\mathrm{C}}\mid p) + |\mathcal{A}|}.
\end{equation}

After construction, taking the leaves of the PST gives the states of the corresponding PSA. However, this simple action may not always admit a valid transition between all states. In this case, leaves must be added to the PST to ensure a complete PSA. Nodes are added to the PST until, for every leaf in the PST, the longest prefix of the leaf exists in the PST. When this condition is true, the leaves of the PST are guaranteed to give a complete PSA~\cite{ron_power_1993}. Transition probabilities between states of the PSA correspond directly to the transition probabilities of the PST found via Eqn.~\ref{eq:trans_prob} and are used to calculate probabilities online.

\subsection{Illustration of $h$-signature Prediction}

\begin{figure}[t!]
    \centering
    \subfloat[Data\label{fig:train_toy}]{\includegraphics[width=0.24\columnwidth]{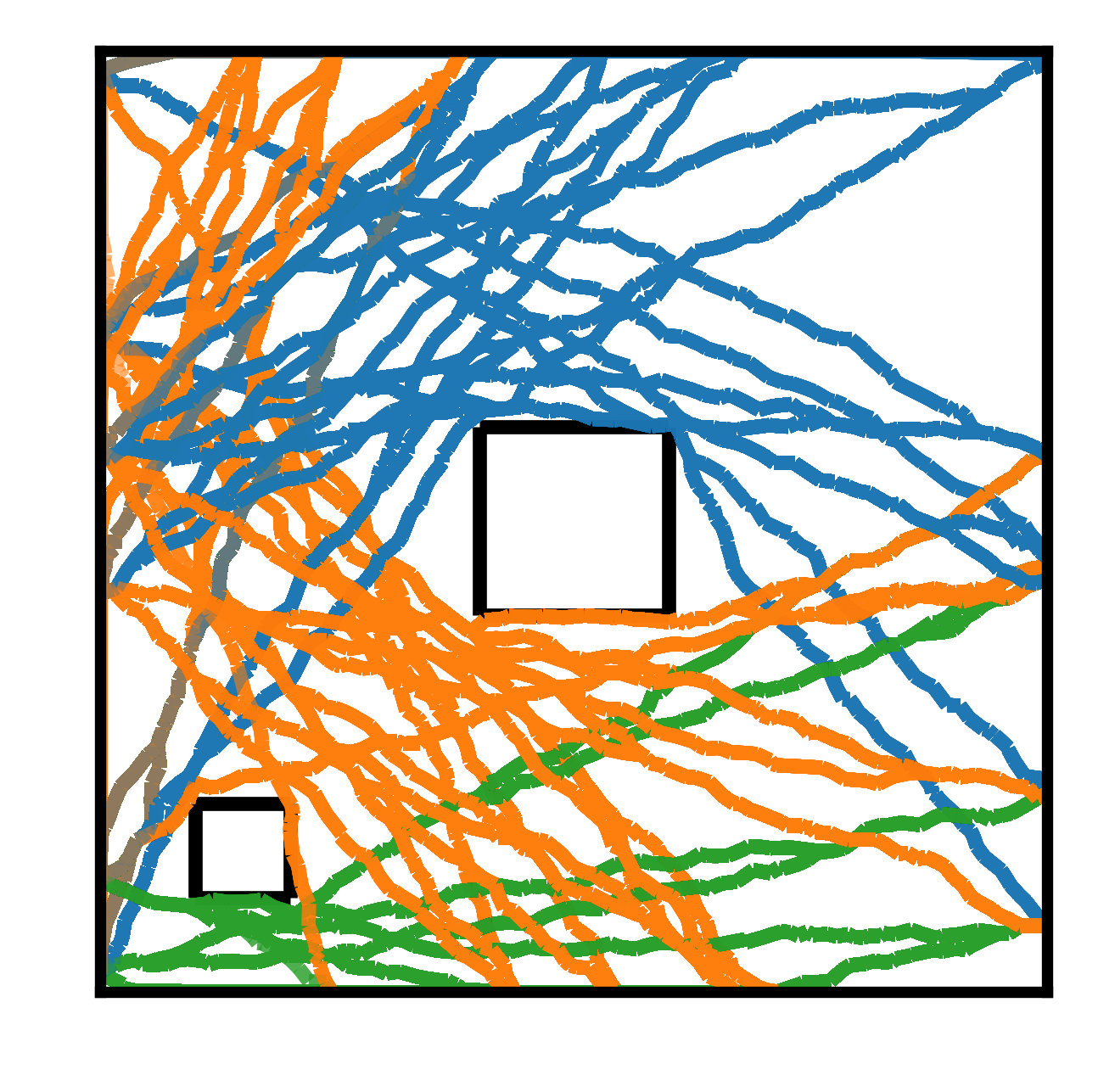}}
    \subfloat[$t=8$\label{fig:toy_vmmp_1}]{\includegraphics[width=0.24\columnwidth]{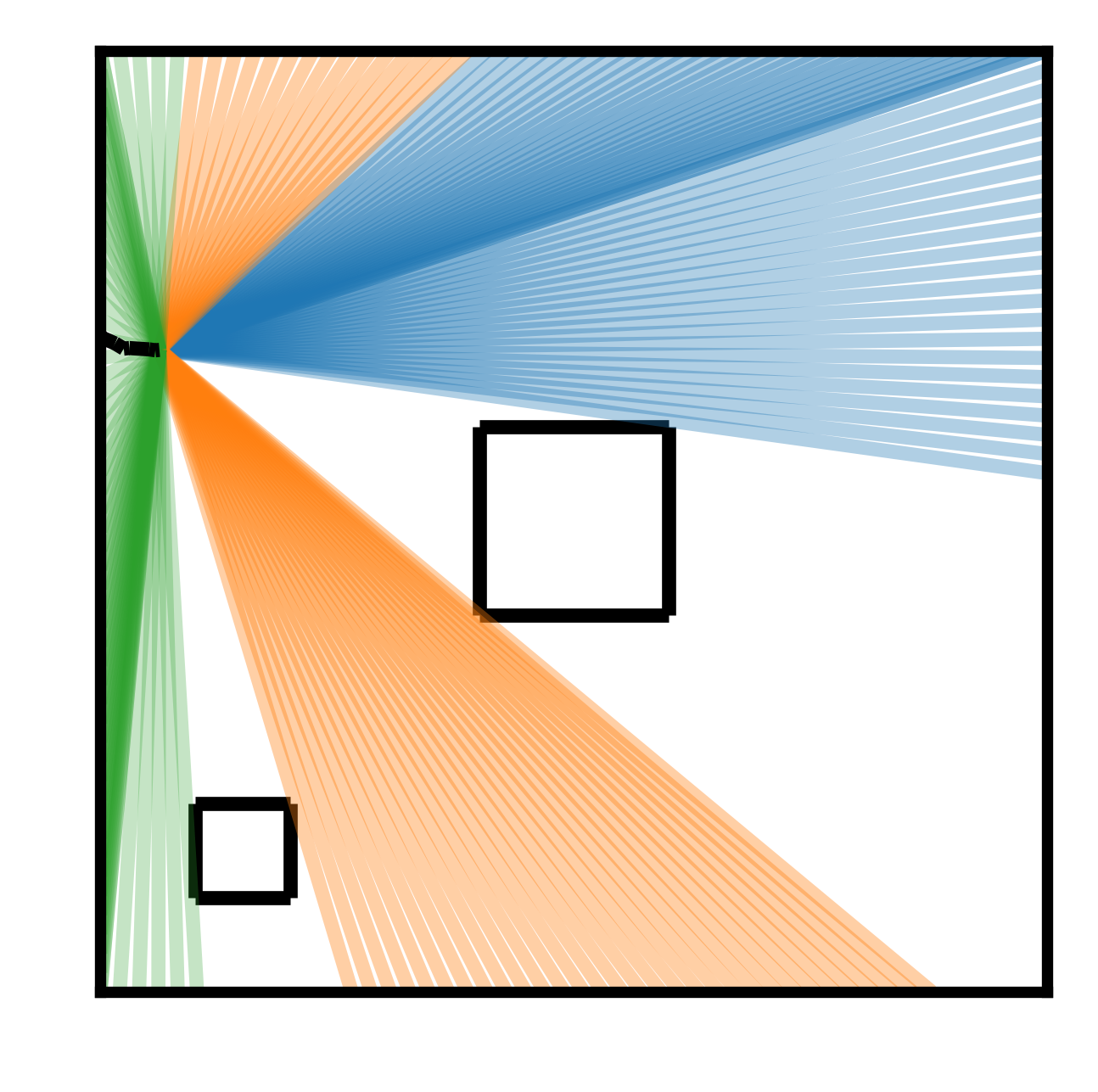}}
    \subfloat[$t=17$]{\includegraphics[width=0.24\columnwidth]{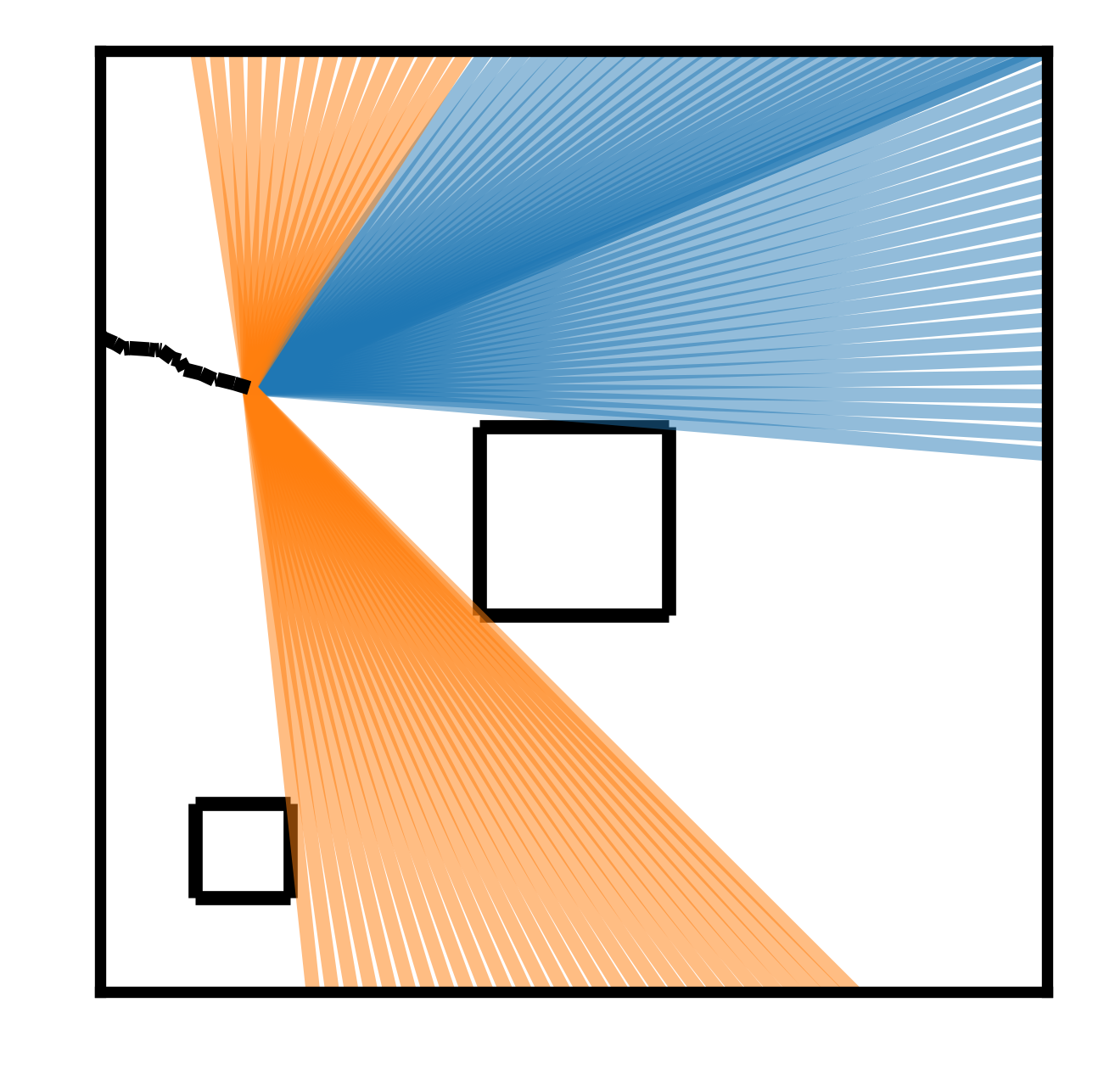}}
    \subfloat[$t=44$\label{fig:toy_vmmp_3}]{\includegraphics[width=0.24\columnwidth]{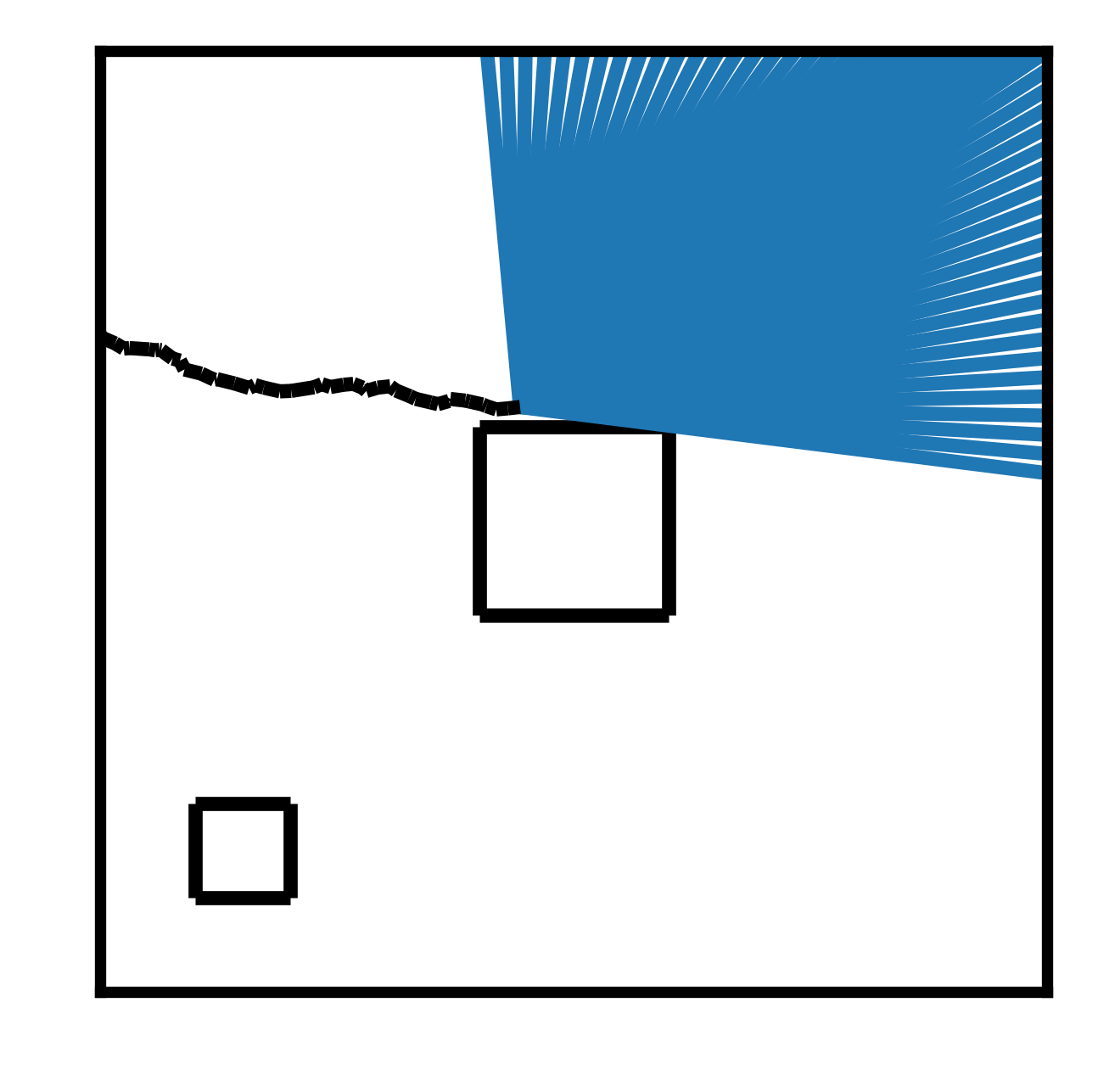}}
    \caption{VOMP $h$-signature prediction for a test trajectory over time. Black trajectory is the observed test trajectory. Coloured regions correspond to straight-line path completions that lie within predicted homotopy classes. Colour coding of homotopy classes is as in the training set. Alpha of the regions reflects the VOMP output probability for that class.} 
    \label{fig:high_level_vmmmp}
    \vspace{-1.5em}
\end{figure}

Here, an illustration of VOMP's high-level motion prediction process is given. A simplistic toy environment and dataset was created, shown in Fig.~\ref{fig:high_level_vmmmp}~\subref{fig:train_toy}. Trajectory data was created by running Dijkstra's algorithm on a graph over the environment to find shortest-distance paths from randomly selected points on the left boundary to random points on all other boundaries. Thus, in this simple dataset all trajectories move from left to right, and the set of possible homotopy classes are denoted by $h$-signatures $\{(), (1), (1,2)\}$.

Prediction of $h$-signatures over time is shown in the following figures Fig.~\ref{fig:high_level_vmmmp}~\subref{fig:toy_vmmp_1}~-~\subref{fig:toy_vmmp_3}. At time $t=8$, the partial $h$-signature is $()$, and the VOMP predicts that all $h$-signature trajectories are possible in the future, with probabilities $P(()) = 0.28$, $P((1)) = 0.37$, $P((1,2)) = 0.35$. At $t=17$ the $()$ class is predicted with probability $0$ now the observed trajectory has passed above the first obstacle, and $P((1)) = 0.52$, $P((1,2)) = 0.48$. In the final time step the VOMP correctly assesses that the trajectory will have full $h$-signature $(1,2)$ with probability 1 as the trajectory has passed the second obstacle. Coloured regions correspond to regions in which one may expect the agent to be in the future, with probabilities indicated by opacity. These regions demonstrate the predictive power of the VOMP output.

\subsection{Low-level Prediction using hierarchical Gaussian Mixture Models}\label{sec:gmm}
We present a hierarchical GMM as an example implementation of the low-level prediction algorithm for retrieving a probability distribution $ P(\mathbf{X} \mid h)$ over the trajectory $\mathbf{X}$ given the final $h$-signature $h$.
To this end, we simply cluster the trajectories from the training dataset into their homotopy classes (i.e. having the same $h$-signature $h$), and fit a GMM for each class,
\begin{equation}\label{eq:prior_gmm_hsig}
    P(\mathbf{X} \mid h) = \sum_{c} w^{(c, h)} \mathcal{N}( \mathbf{X} \mid \mathbf{M}^{ (c, h)}, \Sigma^{(c, h)}).
\end{equation}
Here, for a given $h$-signature $h$, $w^{(c, h)}$, $\mathbf{M}^{ (c, h)}$ and $\Sigma^{(c, h)}$ are the weight, mean and covariance respectively of a component $c \in [1, N_{C}]$.
$\mathcal{N}(\mathbf{X} \mid \mathbf{M}, \Sigma )$ is the multivariate normal distribution over $\mathbf{X}$ with mean vector $\mathbf{M}$ and covariance $\Sigma$.
In doing so, the low-level patterns in the trajectories can be captured by computing the full covariance matrix $\Sigma$, thereby capturing correlations between positions at particular times.


From this distribution, one can derive a fully probabilistic prediction of the trajectory given position measurements $\mathbf{Y}_{\mathrm{obs}}$ and partial $h$-signature $p$. To simplify computation, we treat the partial $h$-signature $p$ as an additional independent measurement to $\mathbf{Y}_{\mathrm{obs}}$.
Then, the partial $h$-signature is conditionally independent of $\mathbf{X}$ given the full $h$-signature $h$, and we can easily condition on the partial $h$-signature as a weighted sum $P(\mathbf{X} \mid p) = \sum_{h} P(\mathbf{X} \mid h) P(h \mid p)$. This effectively only scales the weights of the GMM by $P(h\mid p)$ given by the VOMP.

Subsequently, we can further condition on the actual measurements $\mathbf{Y}_{\mathrm{obs}}$. Following standard methods, 
\begin{equation}
    P(\mathbf{X} \mid p, \mathbf{Y}_{\mathrm{obs}}) = \sum_{c, h} \hat{w}_{\mathrm{obs}}^{(c, h)} \mathcal{N}(\mathbf{X} \mid  \mathbf{\hat{M}}_{\mathrm{obs}}^{(c, h)}, \mathbf{\hat{\Sigma}}_{\mathrm{obs}}^{(c, h)} ).
\end{equation}
Here, the conditional mean and covariance $\mathbf{\hat{M}}_{\mathrm{obs}}^{(c, h)}$, $\mathbf{\hat{\Sigma}}_{\mathrm{obs}}^{(c, h)}$ are calculated in the same manner as standard conditional Gaussian distribution~\cite[Sec.~8.1.3]{matrixcookbook} given measurements.
The conditional weights are calculated as
\begin{equation}
    \hat{w}_{\mathrm{obs}}^{(c, h)} \propto w_{t}^{(c, h)} P(h \mid p) \mathcal{N}( \mathbf{Y}_{\mathrm{obs}} \mid \mathbf{M}^{ (c, h)}, \Sigma^{(c, h)} + \sigma_{\mathbf{Y}}^2 I), \label{eq:conditioning:weights}
\end{equation}
followed by normalisation. The last term is the marginal likelihood of observing $\mathbf{Y}_{t}$ within each mixture component.



\begin{figure*}
    \centering
    \begin{tabular}{c}
        \subfloat[The ATC shopping mall environment and a subset of the data~\cite{atc_dataset}.]{\includegraphics[width=0.285\linewidth]{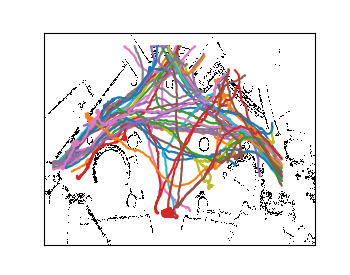}\label{fig:atc_data}}
    \end{tabular}
    \begin{tabular}{c}
        \subfloat[VOMP $t=1$]{\includegraphics[width=0.13\textwidth]{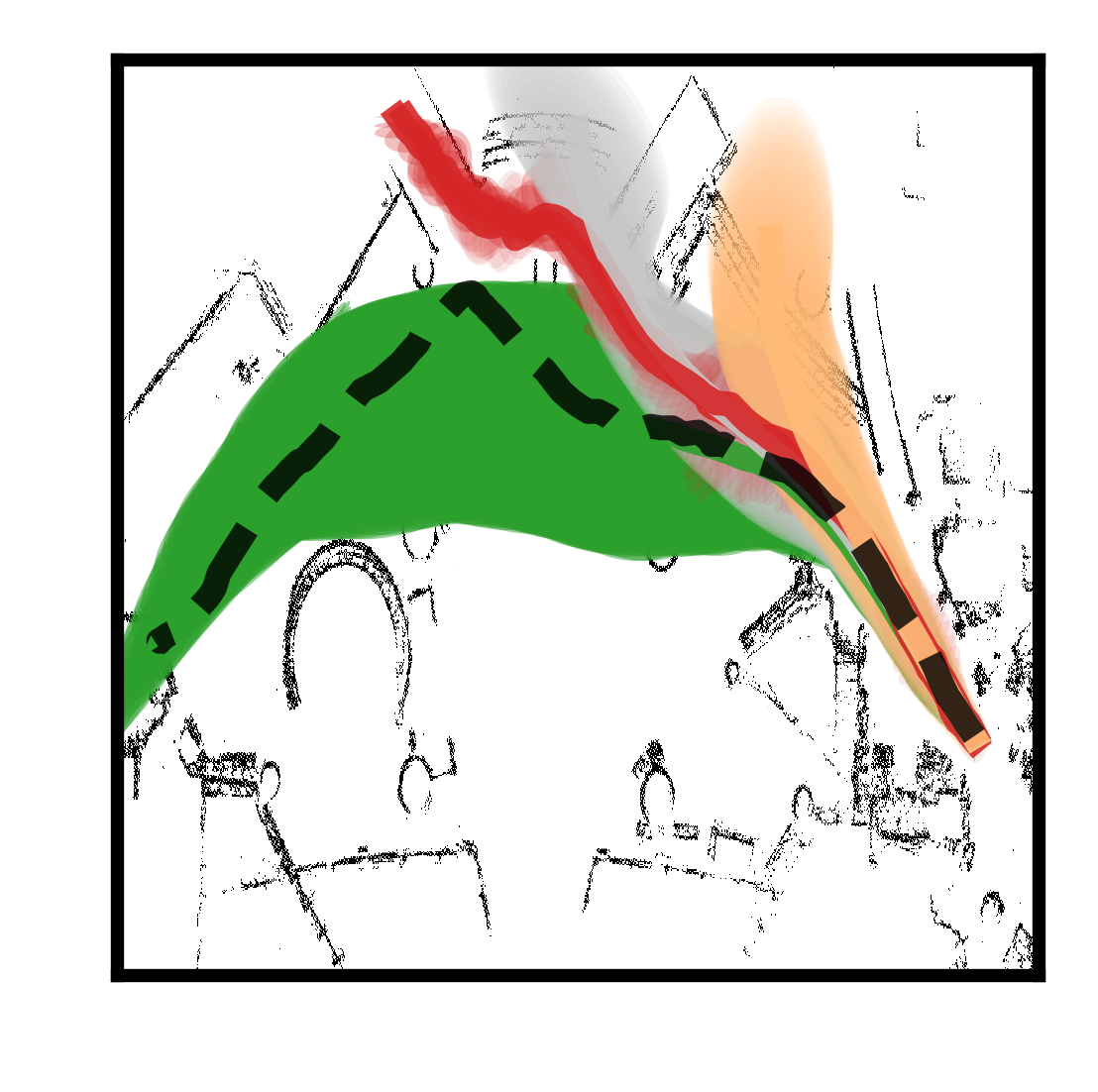}\label{fig:gmmp_1}} 
        \subfloat[VOMP $t=12$]{\includegraphics[width=0.13\textwidth]{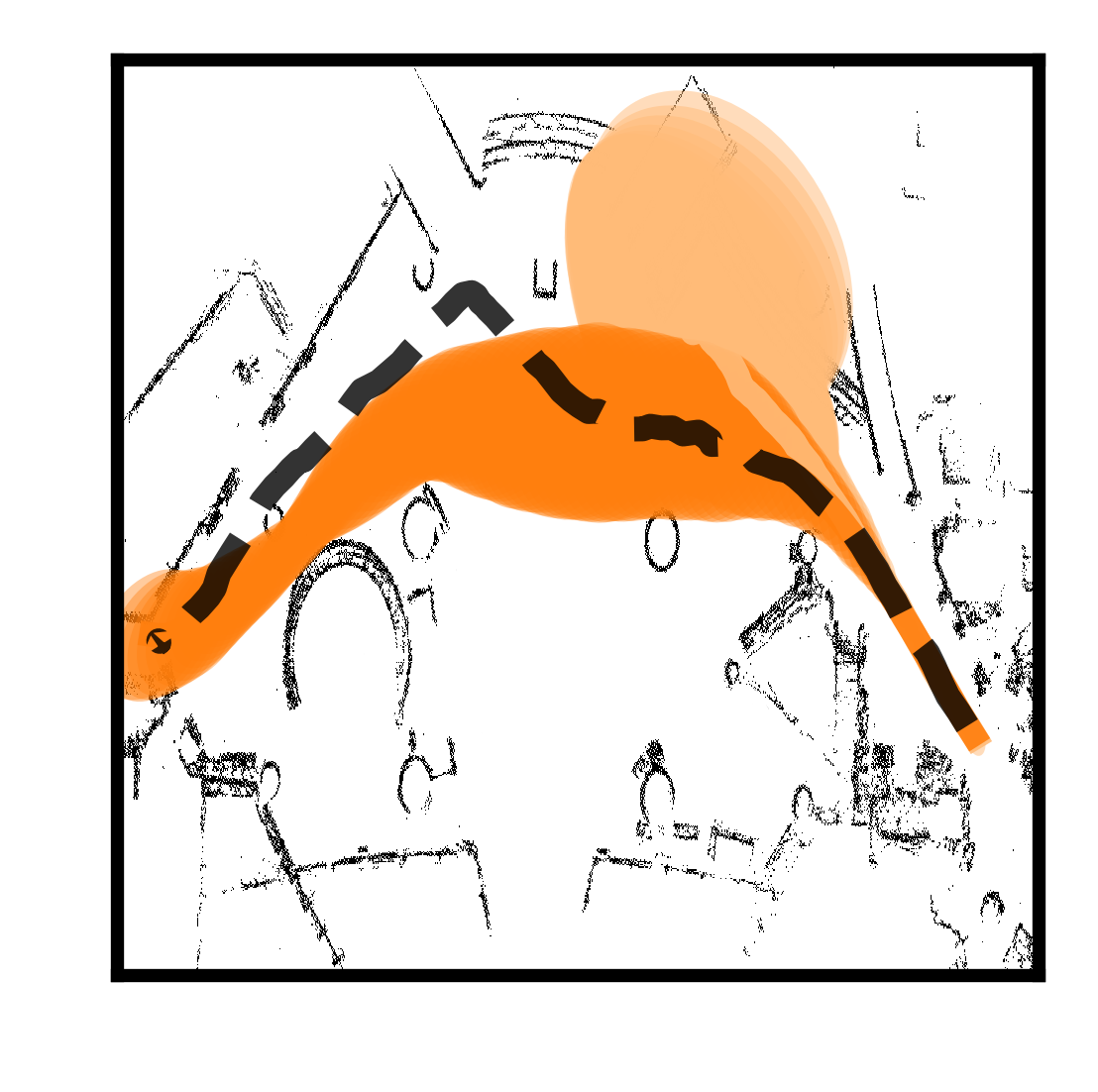}\label{fig:gmmp_2}} 
        \subfloat[VOMP $t=34$]{\includegraphics[width=0.13\textwidth]{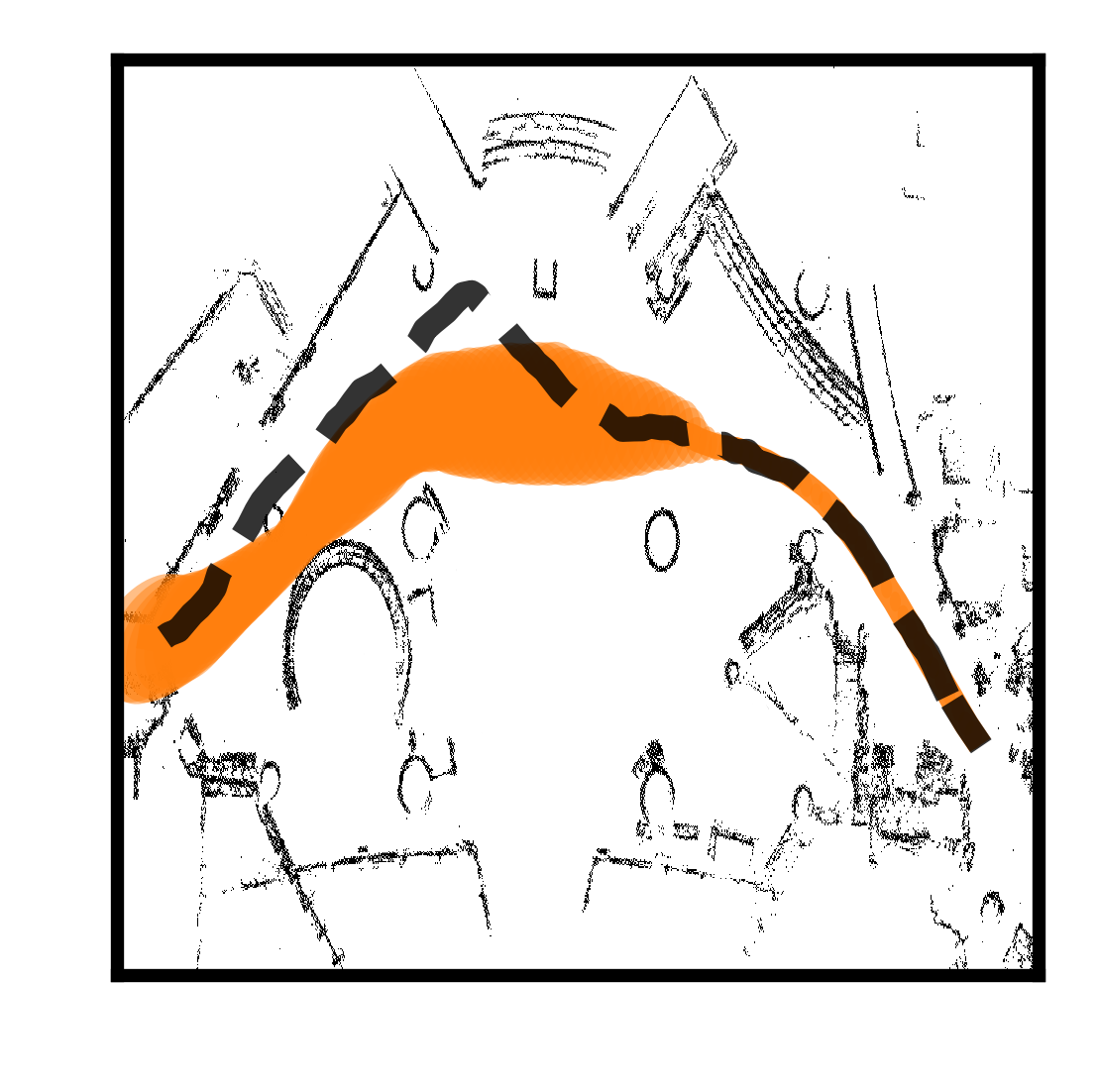}\label{fig:gmmp_3}} 
        \subfloat[VOMP $t=56$]{\includegraphics[width=0.13\textwidth]{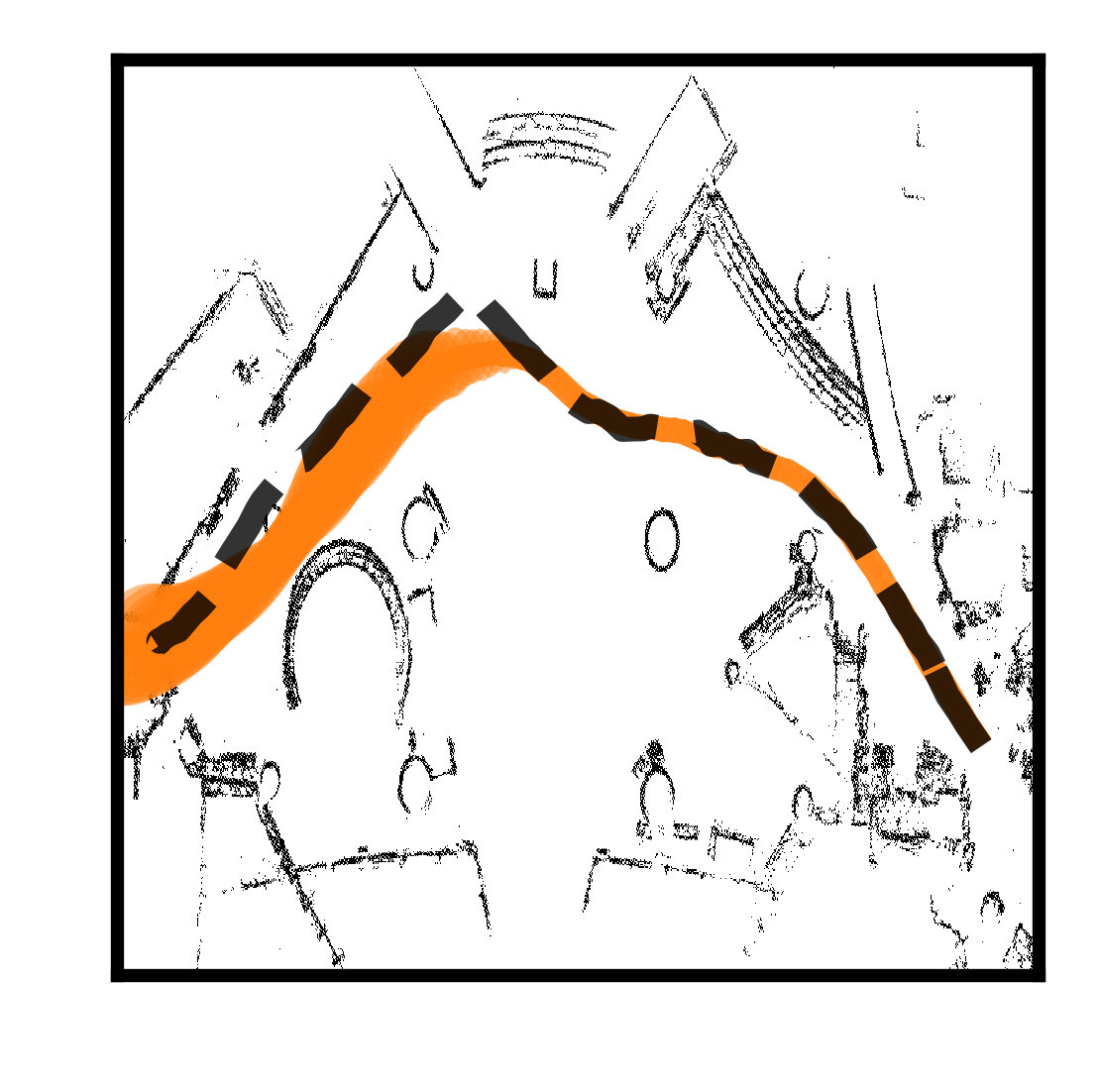}\label{fig:gmmp_4}} 
        \subfloat[VOMP $t=78$]{\includegraphics[width=0.13\textwidth]{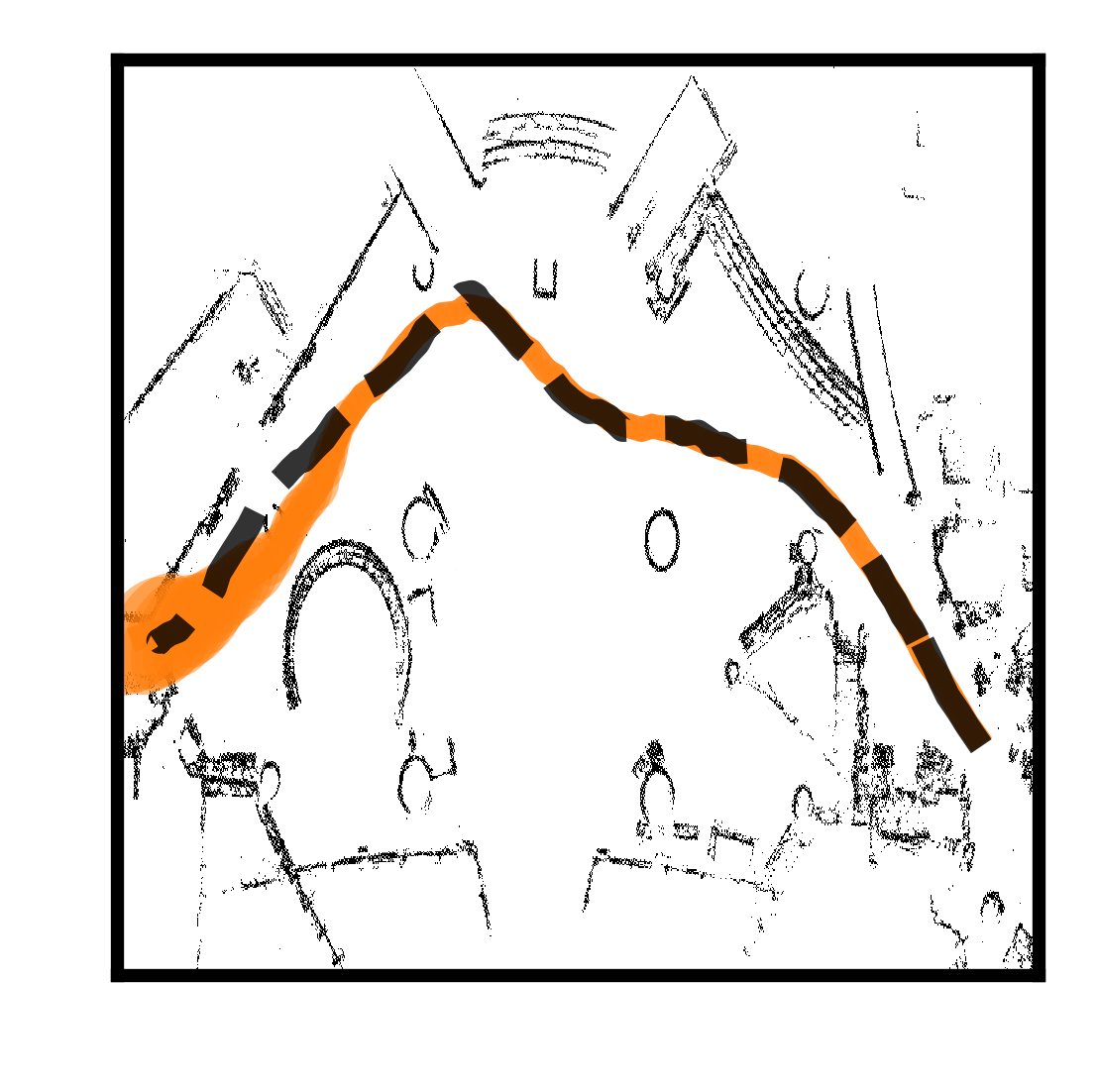}\label{fig:gmmp_5}} \\
        \subfloat[Naive $t=1$]{\includegraphics[width=0.13\textwidth]{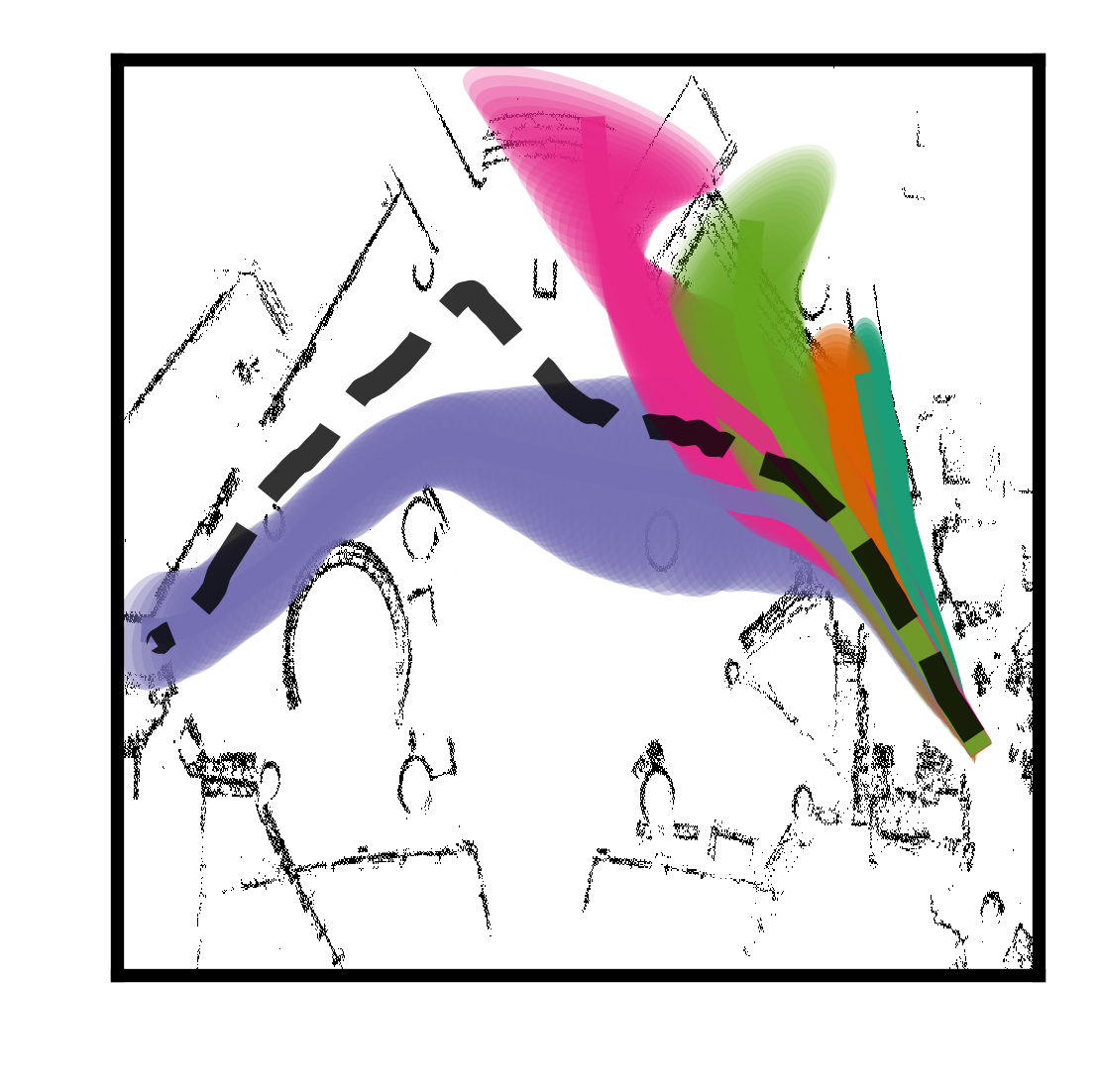}\label{fig:naive_1}} 
        \subfloat[Naive $t=12$]{\includegraphics[width=0.13\textwidth]{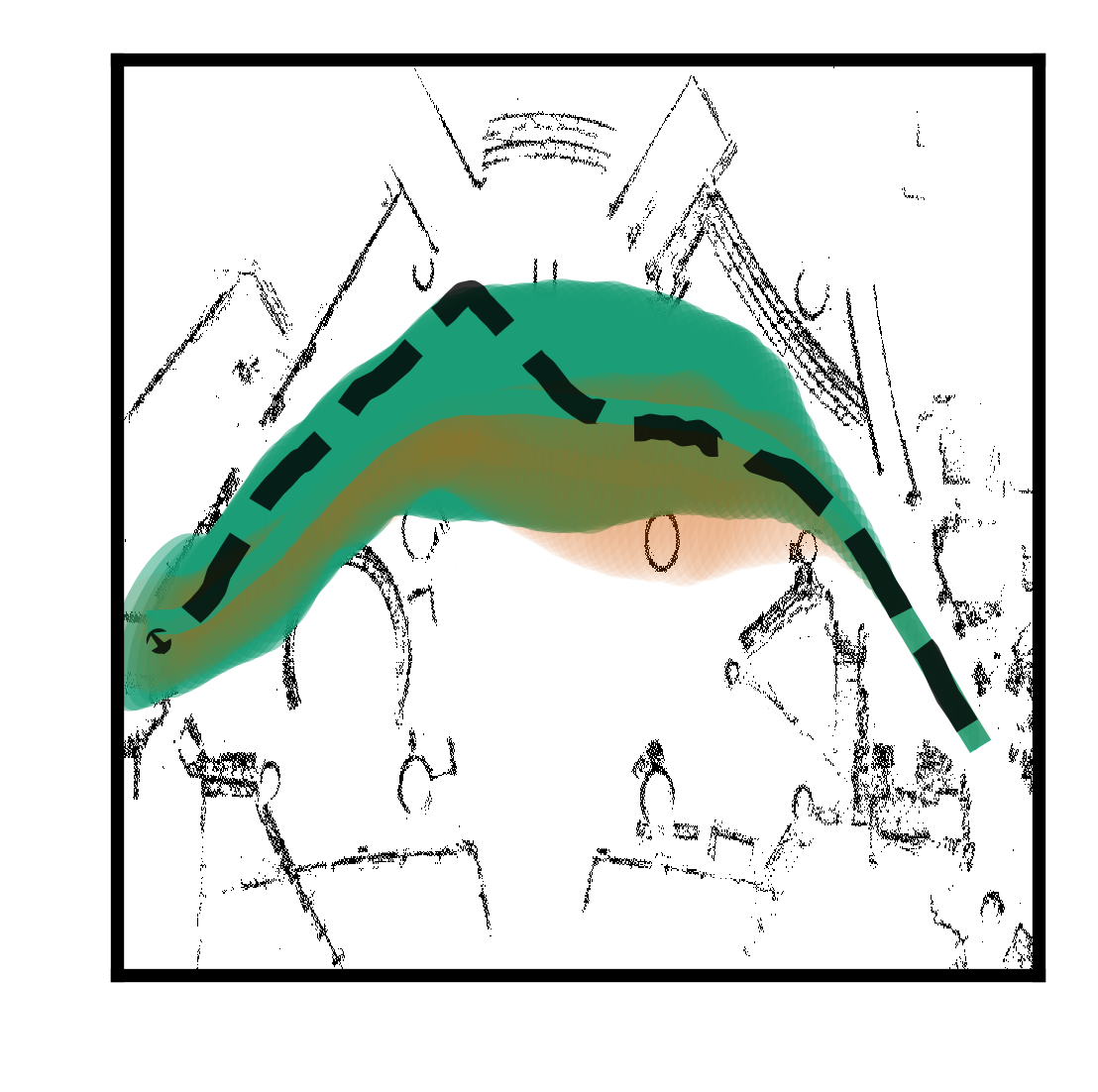}\label{fig:naive_2}} 
        \subfloat[Naive $t=34$]{\includegraphics[width=0.13\textwidth]{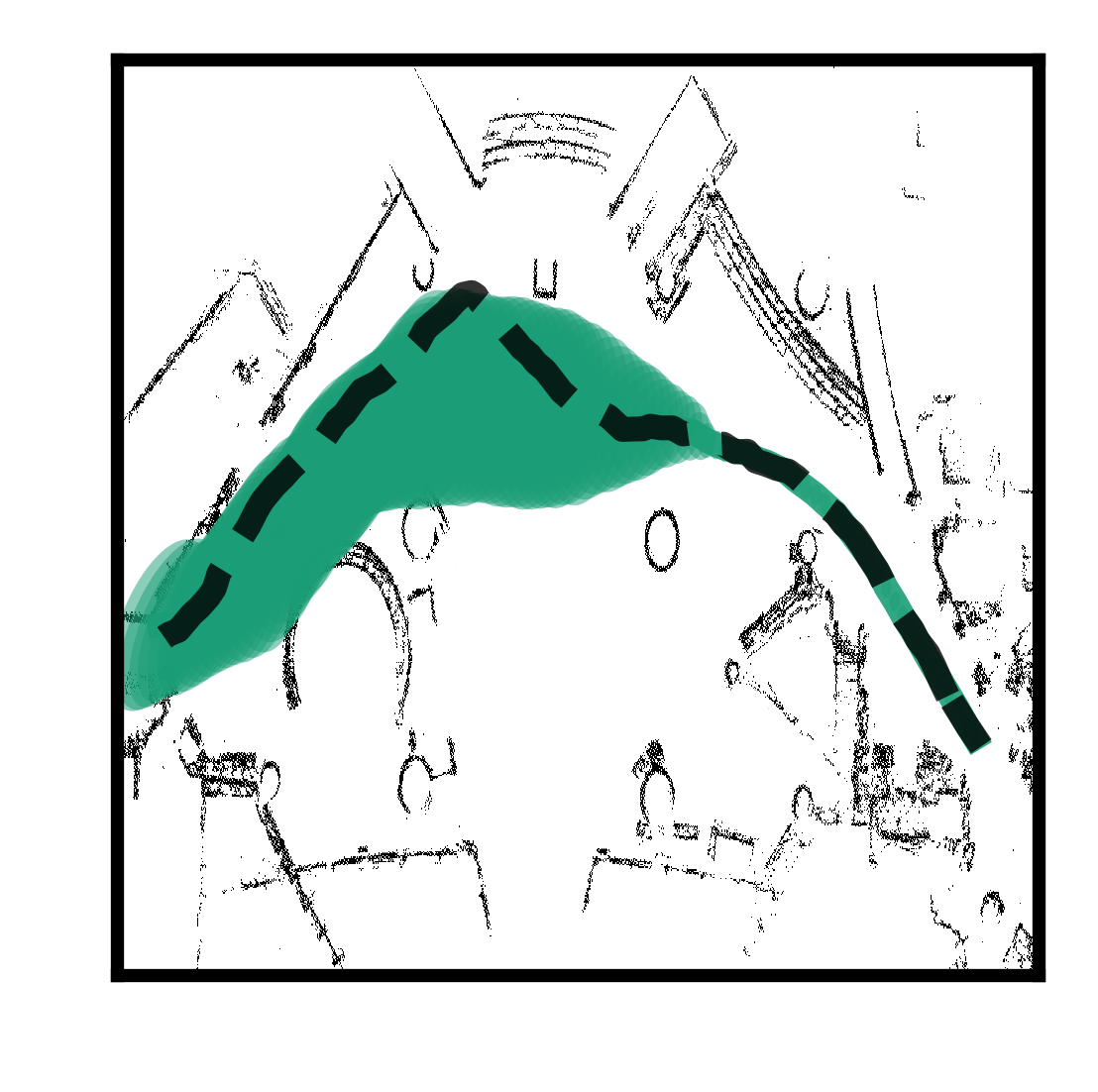}\label{fig:naive_3}} 
        \subfloat[Naive $t=56$]{\includegraphics[width=0.13\textwidth]{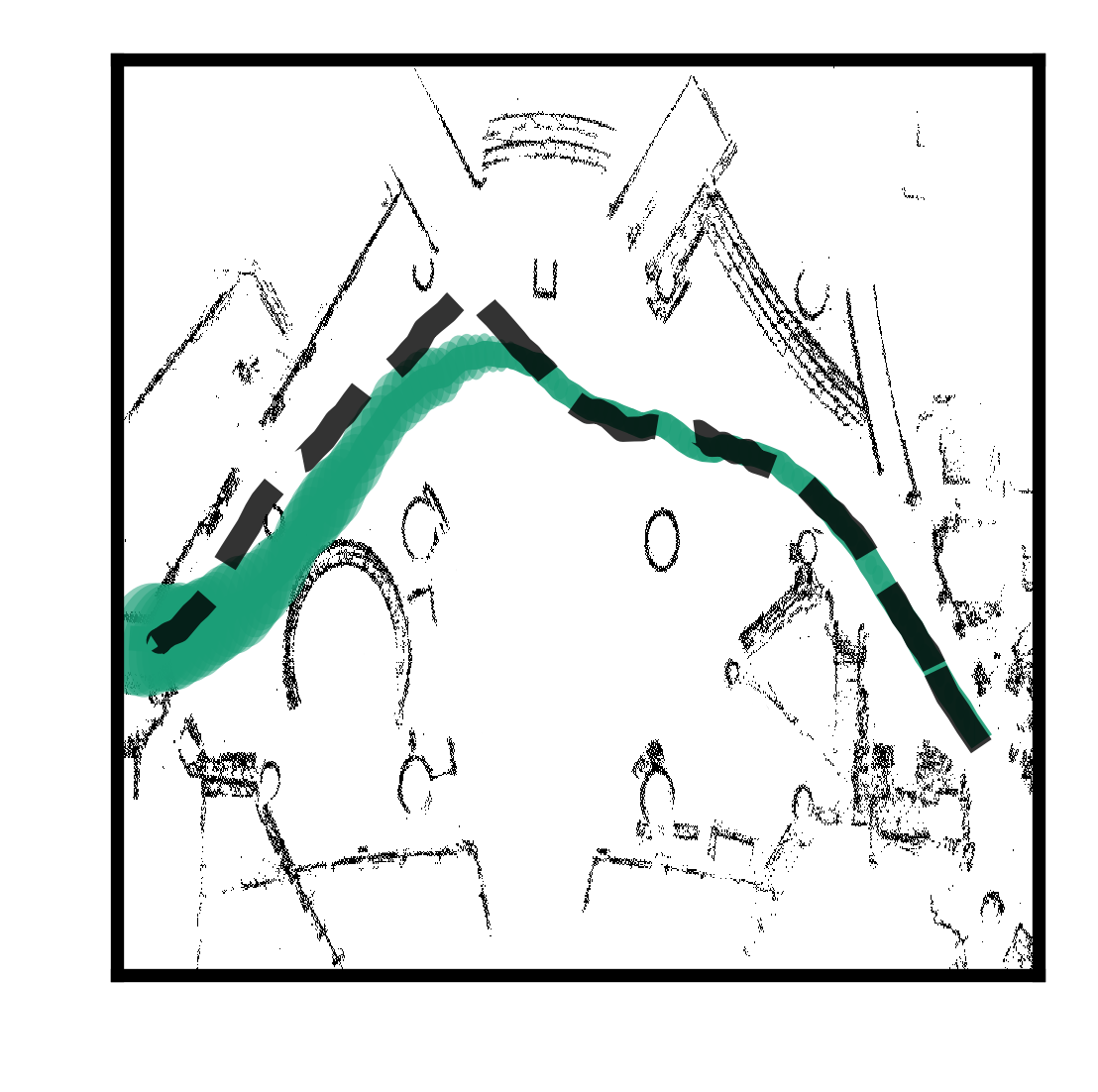}\label{fig:naive_4}} 
        \subfloat[Naive $t=78$]{\includegraphics[width=0.13\textwidth]{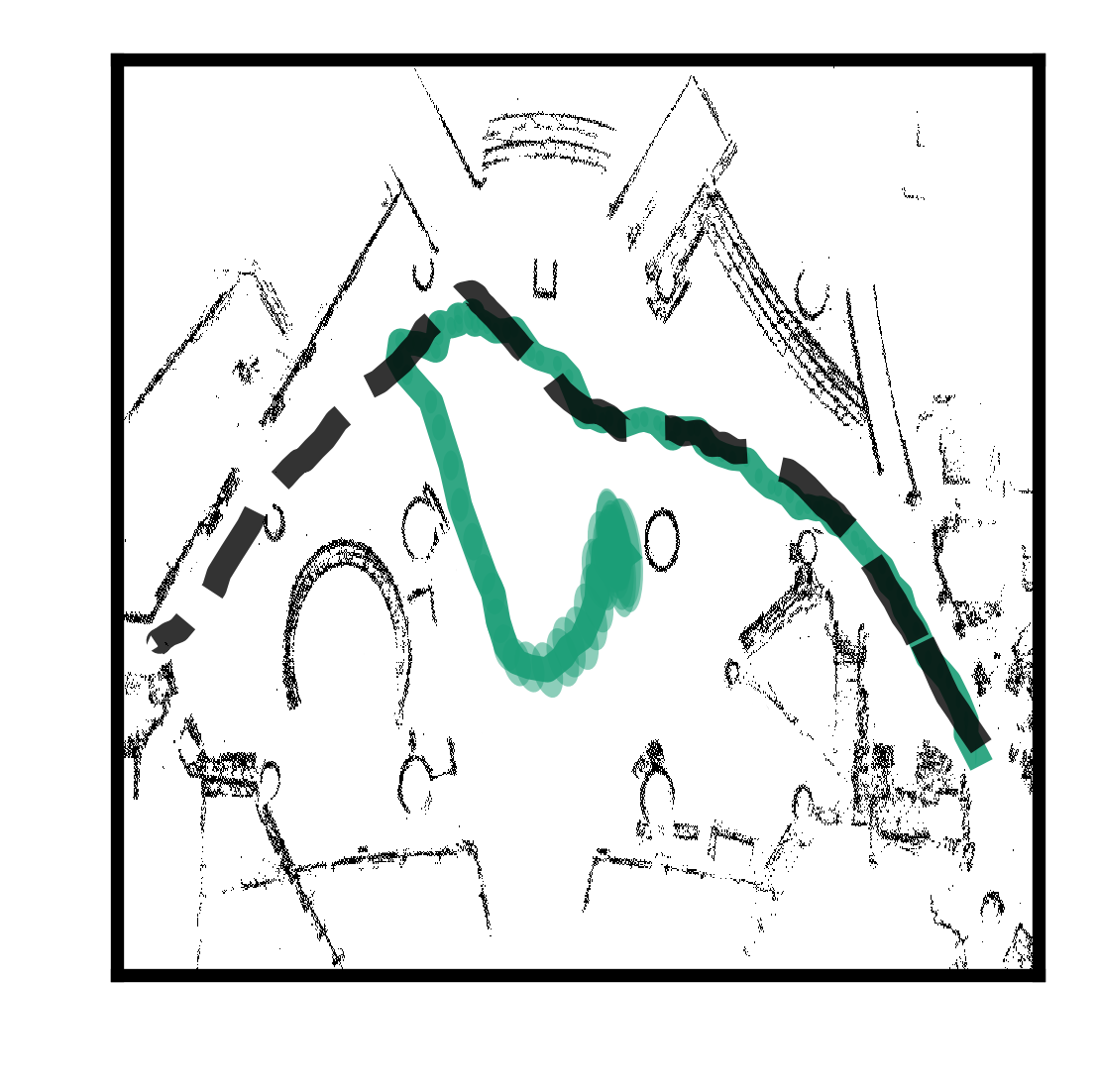}\label{fig:naive_5}}
    \end{tabular}
    \caption{Snapshots of predictions output by a GMM with $h$-signature context given by the VOMP~\protect\subref{fig:gmmp_1}~-~\protect\subref{fig:gmmp_5} and without~\protect\subref{fig:naive_1}~-~\protect\subref{fig:naive_5}. Black dashed trajectory is the ground truth test trajectory. Shaded regions show the variance of around the mean trajectory in solid colour. Transparency is proportional to weight. Colours in~\protect\subref{fig:gmmp_1}~-~\protect\subref{fig:gmmp_5} indicate $h$-signatures, while colours in~\protect\subref{fig:naive_1}~-~\protect\subref{fig:naive_5} indicate mixture components.}
    \label{fig:low_level_gmmp}
    \vspace{-1em}
\end{figure*}

\section{Experimental Results}

\subsection{Experimental Setup}
We experimentally illustrate the benefits of high-level prediction by comparison against a standard GMM without topological knowledge.
We use the ATC shopping mall dataset~\cite{atc_dataset} shown in Fig.~\ref{fig:low_level_gmmp}~\subref{fig:atc_data}.
We retrieved a portion of the dataset containing 17558 trajectories.
We selected 9230 trajectories among the dataset trajectories contained within the area shown in Fig.~\ref{fig:low_level_gmmp}~\subref{fig:atc_data} that satisfy the border crossing assumption.
Among these trajectories, we randomly selected 5538 trajectories to create a training dataset, and 1000 trajectories for testing. 
Trajectories are interpolated over a discrete number of timesteps and used to evaluate the prediction performance of the two pipelines.

Three metrics are used for comparison.
Most immediately, we consider the standard \emph{average displacement error}~(ADE),
which measures the deterministic error in the maximum likelihood prediction. 
This is defined as
\begin{equation}
    \text{ADE} = \frac{1}{T} \sum_{t} || \mathbf{x}_{t} - \mathbf{\hat{M}_{t, obs}}^{ (c^{*}, h^{*})}  ||,
\end{equation}
where $(c^{*}, h^{*})$ denotes the index of the component with the highest weight.  

Since the framework is fully probabilistic, we also need to account for the uncertainty estimates produced. 
We also consider the \emph{average Mahalanobis distance}~(AMD), which accounts for the weights and covariances in the GMM:
\begin{equation}
    \text{AMD} =
            \sum_{c, h} \frac{\hat{w}_{\mathrm{obs}}^{ (c, h) }}{T} \sum_{t} \Theta_{t}^{{ (c, h) }^{\mathrm{T}}} (\mathbf{\hat{\Sigma}_{t, obs}}^{ (c, h ) })^{-1} \Theta_{t}^{ (c, h) },
\end{equation}
where $\Theta_{t}^{ (c, h) } = (\mathbf{x}_{t} - \mathbf{\hat{M}_{t, obs}}^{ (c, h) })$.

Lastly, we would like to characterise the performance of the high-level prediction made.
To do so, we compare the \emph{KL divergence} (KLD) between the posterior GMM weights given the full trajectory and a partial trajectory, defined as
\begin{equation}
    \text{KLD} = \sum_{c, h}  \hat{w}_{T}^{ (c, h) } (\log \hat{w}_{T}^{ (c, h) } - \log \hat{w}_{\mathrm{obs}}^{ (c, h) }).
\end{equation}
For fairness, we use the same number of mixtures for the naive GMM as in the topology-informed GMM. All GMMs used are trained using the implementation in the \textsc{scipy} library, using the default arguments. 



\subsection{Results}

\begin{figure}[t!]
    \centering
    \includegraphics[width=\columnwidth]{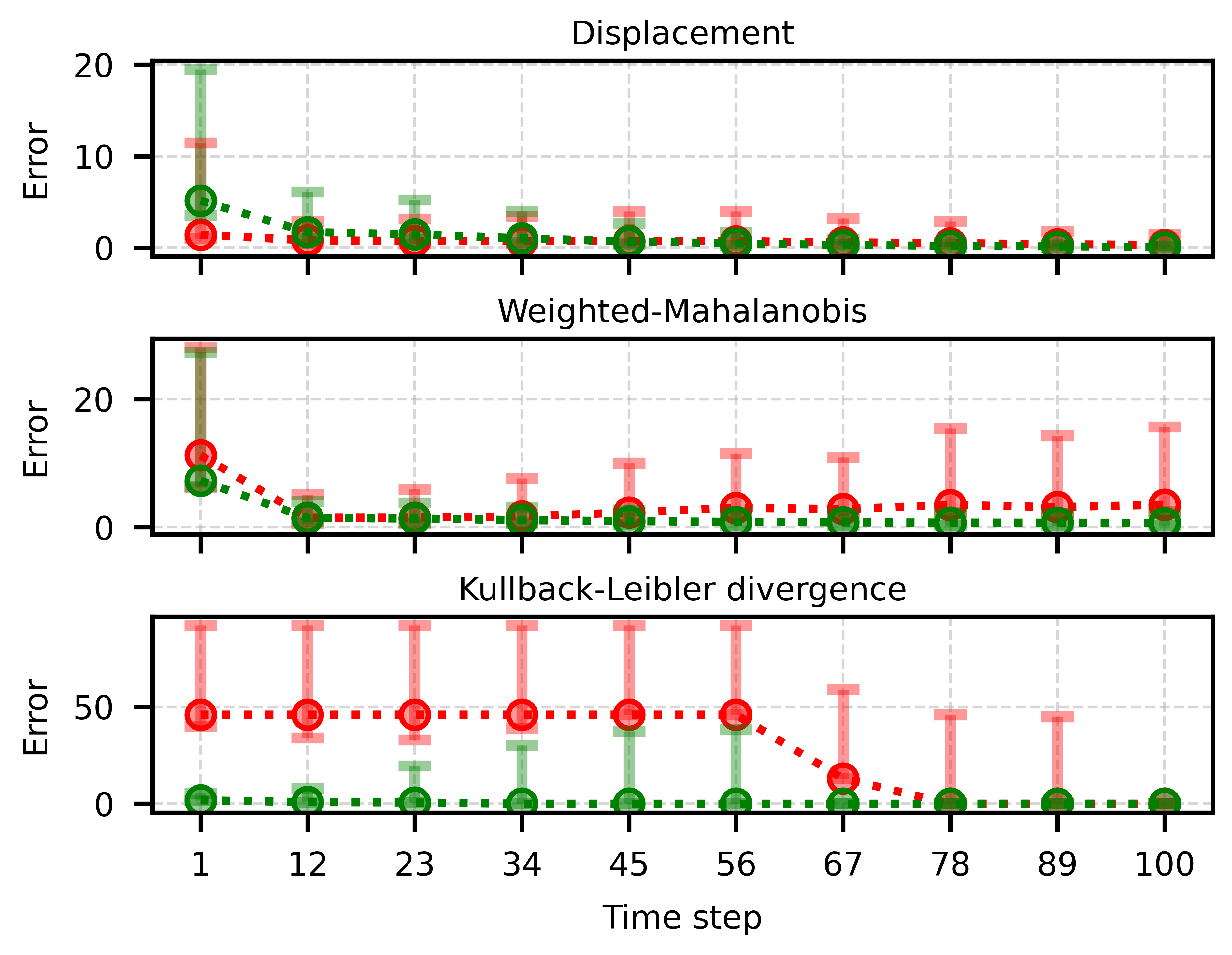}
    \caption{Experimental results. Red: Naive GMM, Green: Our approach. Lower is better. Markers show the median. Error bars show 25\% and 75\% quantiles.} 
    \label{fig:atc_results}
    \vspace{-1.5em}
\end{figure}

Figure~\ref{fig:atc_results} shows the results of comparisons between naive GMM (red) and our topology-informed approach (green) in terms of ADE (top), AMD (middle), and KLD (bottom).
In terms of ADE, our approach outperforms the naive approach after some time, namely showing a $34.8$\% improvement in the median around halfway through the trajectory ($t = 56$), and up to $69.4$\% over time. This increase in performance is because the topology-informed approach gains more information as the target 
crosses obstacles, and this topological information improves the predictive power of the VOMP.

Since the naive GMM was given the same number of components as the topology-informed one calculated based on the number of homotopy classes, it is unsurprising that the naive GMM performs relatively well.
In practice, a major challenge in deploying GMMs is selecting the right number of mixture components. In fact, the naive GMM is initialised with a number directly informed by the number of homotopy classes present in the data, providing a small topological cue to the baseline method.

Nonetheless, the AMD metric shows that our approach performs significantly better than or equal to the naive approach. Specifically, we see a $72.3$\% improvement over the naive halfway through the trajectory, with a maximum of $80.3$\% improvement over time. This implies that the uncertainty predictions from the topology-informed approach are more \emph{consistent} with the actual error compared to the naive approach. This is because our approach pre-clusters trajectories that are `similar', improving the fitness of the GMM model within each cluster.

Further, our approach consistently outperforms in terms of KLD by a great margin, with $100$\% improvement achieved by halfway through the trajectory. This means that early trajectory GMM weights of our approach better represented the final GMM weights. Viewing GMM components as an alternative description of high-level motion to $h$-signatures, this implies that $h$-signatures are indeed better descriptors.

Figure~\ref{fig:low_level_gmmp} shows the behaviour of our approach and baseline GMM.
Early on, we see both methods give highest weight to components that reasonably predict the general motion of the trajectory. However, the naive approach begins to produce poor predictions in an area of the environment where many training paths diverge. This `crossroads' is visible in the sample data plotted in Fig.~\ref{fig:low_level_gmmp}~\subref{fig:atc_data}. In comparison, our approach continues to correctly predict the same $h$-signature at this crossroads. This demonstrates the power of topological information in imbuing low-level predictions with robustness to deviations in trajectories.

\section{Conclusion}
We presented a framework for trajectory prediction using homotopy classes. 
The core of the framework is the notion of partial $h$-signature, which allows prediction of the full trajectory at both abstract and geometric levels. 
Under this framework, we presented VOMP and hierarchical GMMs as minimal implementations of the high-level and low-level prediction components, the combination of which was shown to outperform baselines without topological knowledge.

Our framework opens many exciting opportunities for future work. 
The high-level and low-level components are modular, and can be replaced with many other algorithms of varying sophistication and efficiency.
Further, we expect the saliency and sparseness of $h$-signatures will be instrumental when directly used in planning problems such as target search in cluttered environments~\cite{wakulicz_active_2021}. 
We believe similar topological features can be defined in other application domains such as marine buoys~\cite{argo,floats2012}. 

\bibliographystyle{IEEEtran}
\bibliography{bib}

\end{document}